\documentclass[lettersize,journal]{IEEEtran}
\usepackage{amsmath,amsfonts}
\usepackage{algorithmic}
\usepackage{algorithm}
\usepackage{array}
\usepackage[caption=false,font=normalsize,labelfont=sf,textfont=sf]{subfig}
\usepackage{textcomp}
\usepackage{stfloats}
\usepackage{url}
\usepackage{verbatim}
\usepackage{graphicx}
\usepackage{cite}

\usepackage{hyperref}
\usepackage{amssymb}

\usepackage{threeparttable}
\usepackage{booktabs}

\usepackage{picins}
\usepackage{xcolor, soul}
\usepackage{url}

\begin{document}

\title{Non-Autoregressive Diffusion-based Temporal Point Processes for Continuous-Time Long-Term Event Prediction}

\author{Wang-Tao Zhou, Zhao Kang, Ling Tian}


\markboth{Preprint,~Vol.~XX, No.~X, XX~XXXX}%
{Shell \MakeLowercase{\textit{et al.}}: A Sample Article Using IEEEtran.cls for IEEE Journals}


\maketitle

\begin{abstract}
Continuous-time long-term event prediction plays an important role in many application scenarios. Most existing works rely on autoregressive frameworks to predict event sequences, which suffer from error accumulation, thus compromising prediction quality. Inspired by the success of denoising diffusion probabilistic models, we propose a diffusion-based non-autoregressive temporal point process model for long-term event prediction in continuous time. Instead of generating one event at a time in an autoregressive way, our model predicts the future event sequence entirely as a whole. In order to perform diffusion processes on event sequences, we develop a bidirectional map between target event sequences and the Euclidean vector space. Furthermore, we design a novel denoising network to capture both sequential and contextual features for better sample quality. Extensive experiments are conducted to prove the superiority of our proposed model over state-of-the-art methods on long-term event prediction in continuous time. To the best of our knowledge, this is the first work to apply diffusion methods to long-term event prediction problems.
\end{abstract}

\begin{IEEEkeywords}
Temporal Point Processes, diffusion, long-term, event prediction, non-autoregressive.
\end{IEEEkeywords}

\section{Introduction}
Event prediction is an essential task in many application areas, such as finance~\mbox{\cite{finance}}, traffic~\mbox{\cite{traffic}}, healthcare~\mbox{\cite{healthcare}}, recommendation~\mbox{\cite{recommendation}}, etc. Enhancing the accuracy and reliability of event prediction systems can help to anticipate potential opportunities and risks in the future, in order to maximize the benefits of human.

Temporal Point Processes (TPPs)~\mbox{\cite{tppintro}} are a very useful tool for modeling discrete events in continuous time, and have been extensively studied in recent years. The Hawkes process~\mbox{\cite{hawkes}} is an early work on TPP that uses simple parametric forms to model self-exciting event patterns. However, such a simple parameterisation is not sufficient to capture complex temporal dependencies. Thus, some recent works have focused on neural TPP models, which apply deep neural networks to model the probability distribution of event occurrences.~\mbox{\cite{rmtpp,nhp,lstmtpp}} adopt different types of recurrent neural networks (RNN) to encode history events and predict the intensity function of the next event using the hidden representation. Some variants improve neural TPP models by applying techniques like fully connected neural networks~\mbox{\cite{FullyNN}}, Transformers~\mbox{\cite{SAHP,THP,Fourier}}, mixture distributions~\mbox{\cite{lognormmix}}, basis functions~\mbox{\cite{unipoint}}, Convolutional Neural Networks~\mbox{\cite{ctpp}}, etc. These works mainly focus on boosting the accuracy of one-step-ahead probabilistic prediction.

\begin{figure*}[!t]
\centering
\subfloat[Autoregressive]{\includegraphics[width=3.5in]{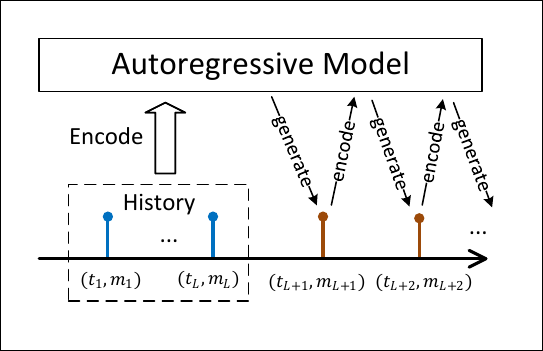}
\label{autoregressive}}
\hfil
\subfloat[Non-autoregressive]{\includegraphics[width=3.5in]{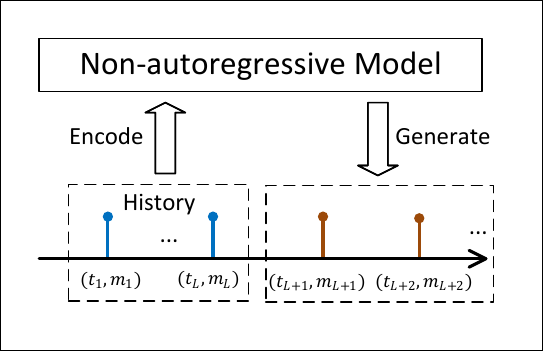}
\label{non-autoregressive}}
\caption{The model-data interaction of autoregressive and non-autoregressive TPP models. Autoregressive models generate an event at a time, which is then fed back to the model to form a new context for the next event. Non-autoregressive models generate multi-step events as a whole, which alleviates error accumulation.}
\label{fig:interaction}
\end{figure*}

Long-term event prediction in continuous time enables us to detect risks or opportunities in the distant future. Most existing TPP methods are autoregressive, and can only model the probability distribution of the next event. Thus, during the process of multi-step prediction, they generate only one event at a time and feed it back to the network to obtain the context for the next step. Fig.~\mbox{\ref{autoregressive}} shows the model-data interaction of autoregressive models. The overall accuracy of the predicted sequences could be compromised due to error accumulation. This calls for the need of non-autoregressive long-term TPP models, which generate multi-step events altogether as a whole (see Fig.~\mbox{\ref{non-autoregressive}}).~\mbox{\cite{long}} is a recent work on long-horizon TPP modelling, which combines a normal autoregressive model with a count model, predicting the number of events that fall into different time bins.~\mbox{\cite{hypro}} trains a separate discriminator to score the event sequences generated by the autoregressive generator. These two works share the idea of using an external structure to alleviate the effect of error accumulation. Although they are generally non-autoregressive, an autoregressive model is internally applied for sequence generation, and thus error accumulation persists. Furthermore, adding additional external structures complicates training and sampling. To this end, we propose a completely non-autoregressive long-term TPP model to tackle this issue.

Generative models have been a research hotspot in recent years, including Variational Auto Encoders (VAEs)~\mbox{\cite{vae}}, Denoising Diffusion Probabilistic Models (DDPMs)~\mbox{\cite{Ho}}, Normalizing Flows~\mbox{\cite{nf}}, and Generative Adversarial Networks (GANs)~\mbox{\cite{GAN}}. Inspired by the success of generative models in other areas, generative TPP models have also been proposed in recent years, such as VAE-based TPP models~\mbox{\cite{vaetpp1,vaetpp2,GNTPP}}, GAN-based TPP models~\mbox{\cite{wasserstein,GNTPP}}, flow-based TPP models~\mbox{\cite{tritpp,GNTPP,lognormmix}}, diffusion-based TPP model~\mbox{\cite{GNTPP}}, etc. Most existing generative TPP models are autoregressive, which are not good at long-term prediction.~\mbox{\cite{wasserstein,tritpp}} focus TPP sequence generation but do not consider event marks and history dependency. DDPMs have achieved great success in image generation~\mbox{\cite{Ho}}, audio synthesis~\mbox{\cite{diffwave}}, language models~\mbox{\cite{diffusion-LM}}, etc. However, DDPM-based long-term TPP modelling has not been researched. Therefore, in this work, we propose applying the idea of DDPMs to history-conditioned TPP sequence generation, which fulfills the goal of long-term event prediction in continuous time.

This work develops a non-autoregressive Diffusion-based Long-term Temporal Point Process (DLTPP).  Our contribution is three-fold:
\begin{itemize}
    \item Inspired by~\mbox{\cite{diffusion-LM}}, we develop a way to conduct the diffusion process on TPP event sequences instead of images. By adding a probabilistic embedding layer that maps event sequences to continuous vectors, we can then perform vanilla DDPMs in the continuous Euclidean space.
    \item To capture both sequential and contextual features of event sequences, we propose a novel denoising network framework, which combines an RNN with a convolutional residual network.
    \item Extensive experiments are performed on synthetic and real-world datasets. 
\end{itemize}

\section{Related Works}
TPP models~\mbox{\cite{tppintro}} are handy tools for event prediction in continuous time. Traditional TPP models, such as Hawkes' processes~\mbox{\cite{hawkes}} and self-correcting processes~\mbox{\cite{self-correct}}, use simple parametric function forms to model conditional event intensities. In recent years, many neural TPP models have been proposed to enhance modelling flexibility. Neural TPP models generally comply with the encoder-decoder structure. RMTPP~\mbox{\cite{rmtpp}} and NHP~\mbox{\cite{nhp}} are two early examples, which adopt different types of continuous-time RNNs to encode history events and use a decoder network to generate the prediction of the intensity function. Some later works enhance the modelling performance by designing different types of encoders. SAHP~\mbox{\cite{SAHP}}, THP~\mbox{\cite{THP}} and AttNHP~\mbox{\cite{attnhp}} adopt the Transformer structure for the history encoder to better capture long-term dependencies. CTPP~\mbox{\cite{ctpp}} combines the RNNs with a continuous-time convolutional network to incorporate global and local event contexts. Some other works focus on the decoder side of the framework. UNIPoint~\mbox{\cite{unipoint}} formulates the intensity function as the combination of multiple base functions. LogNormMix~\mbox{\cite{lognormmix}} adopts the approach of mixture distributions  to achieve intensity-free TPP modelling. \mbox{\cite{vaetpp1,vaetpp2,GNTPP}} to use probabilistic generative modules as the event decoders, including methods like VAEs, normalizing flows, GANs, etc. \mbox{\cite{reinforcement}} proposes a reinforcement learning approach for TPP models. \mbox{\cite{initiator,contrastive}} learn event sequences in a noise-contrastive way. Models mentioned above are only able to perform one-step-ahead prediction. Although we can use these models for multi-step prediction in an autoregressive way, the accuracy could be compromised due to error accumulation. 

Some other works propose to solve the problem of long-term event prediction.~\mbox{\cite{wasserstein}} learns TPP event sequence generation models using Wasserstein GAN as the backbone, where a discriminator network is trained to score the quality of the generated sequences.~\mbox{\cite{tritpp}} presents a probabilistic sequence generation model based on normalizing flows. These models predict the target sequence as a whole rather than only predicting one event at a time. However, they do not condition the prediction on history events. \mbox{\cite{dualtpp,hypro}} are two recently published works on history-conditioned long-term TPP sequence prediction, which apply macro statistics or discriminative networks to re-rank the generated sequences, enhancing their sample quality. However, apart from the re-ranking, these models still rely on autoregressive sequence generators, which again suffer from error accumulation problem. To resolve this issue, we propose a non-autoregressive long-term TPP sequence generation approach to further improve the prediction performance.

DDPMs~\mbox{\cite{Ho,diffusion2,diffusion3}} were originally proposed for image generation, and they can produce synthetic images with higher quality compared to previous models. DDPMs have also been used to generate non-image data, such as audio synthesis~\mbox{\cite{diffwave}}, natural language generation~\mbox{\cite{diffusion-LM}}. Diffusion methods have been employed for temporal prediction in recent years. TimeGrad~\mbox{\cite{TimeGrad}} is an autoregressive multivariate time series forecasting model. D\textsuperscript{3}VAE \mbox{\cite{D3VAE}} uses a bidirectional auto-encoder (BVAE) together with denoising score matching to clear the noise.~\mbox{\cite{non-autoregressive}} proposes a non-autoregressive multi-step time series forecasting model, which solves a similar issue as ours, but an event sequence is comprised of discrete event marks and non-decreasing arrival times, which is different from normal time series data. DDPMs have already been applied to TPP modeling. \mbox{\cite{spatio-temporal-diffusion}} is a spatio-temporal point process model based on diffusion models with a spatio-temporal co-attention denoising network. However, this work still focuses on one-step-ahead prediction and does not support long-term prediction. Therefore, in this work, we develop a non-autoregressive long-term TPP sequence generation model with a novel denoising structure.

\section{Preliminary}
\label{preliminary}
\subsection{Denoising Diffusion Probabilistic Models}
\label{ddpm}
Denoising Diffusion Probabilistic Models (DDPMs)~\mbox{\cite{Ho}} generate target samples by taking a standard Gaussian variable and gradually removing the noise step by step using a denoising network. The denoising network can be trained by adding random noise to the real data and forcing the network to predict the noise we add. Suppose we have data $ \boldsymbol{x}^0 $, we gradually add Gaussian noise to it using a Markov chain:
\begin{equation}
\label{forward1}
    q(\boldsymbol{x}^{1:N}|\boldsymbol{x}^0)=\prod_{n=1}^N q(\boldsymbol{x}^n|\boldsymbol{x}^{n-1})
\end{equation}
where $ N $ is the number of maximum diffusion steps. The noise-adding steps are defined as follows:
\begin{equation}
\label{forward2}
    q(\boldsymbol{x}^n|\boldsymbol{x}^{n-1}):=\mathcal{N}(\boldsymbol{x}^n;\sqrt{1-\beta_n}\boldsymbol{x}^{n-1},\beta_nI)
\end{equation}
The whole noise-adding procedure is known as the forward process, which is not trainable, having hyperparameters $ \beta_1, ..., \beta_N \in (0, 1) $. According to~\mbox{\cite{Ho}}, the forward process can be accelerated by pre-computing parameters $ \alpha_n=1-\beta_n $ and $ \Bar{\alpha}_n=\prod_{r=0}^n \alpha_n $. In this case, the corrupted sample at the $ n $-th level can be obtained in closed form using the reparameterization trick:
\begin{equation}
\label{forward}
    \boldsymbol{x}^n=\sqrt{\Bar{\alpha}_n}\boldsymbol{x}^0+\sqrt{1-\Bar{\alpha}_n}\boldsymbol{\epsilon}
\end{equation}
where $ \boldsymbol{\epsilon}\sim \mathcal{N}(0,1) $ is the standard Gaussian noise.

The denoising process, named the reverse process, is also defined as a Markov chain:
\begin{equation}
\label{eq:reverse}
    p_\theta(\boldsymbol{x}^{0:N}):=p(\boldsymbol{x}^N)\prod_{n=N}^1 p_\theta(\boldsymbol{x}^{n-1}|\boldsymbol{x}^{n})
\end{equation}
The denoising probability is a Gaussian distribution whose mean $ \boldsymbol{\mu}_\theta $ and variance $ \boldsymbol{\Sigma}_\theta $ are parameterized with a neural network:
\begin{equation}
    p_\theta(\boldsymbol{x}^{n-1}|\boldsymbol{x}^{n})=\mathcal{N}\Bigl(\boldsymbol{x}^{n-1}; \boldsymbol{\mu}_\theta(\boldsymbol{x}^{n},n),\boldsymbol{\Sigma}_\theta(\boldsymbol{x}^{n},n)\Bigr)
\end{equation}
This network is optimized to eliminate the Gaussian noise added at the corresponding step $ n $. Similar to VAEs, DDPMs are trained by maximising the Evidence Lower Bound (ELBO):
\begin{equation}
    \mathrm{ELBO}(\theta)=\mathbb{E}_{q(\boldsymbol{x}^{0:N})}\Bigl[\log p_\theta(\boldsymbol{x}^{0:N})-\log q(\boldsymbol{x}^{1:N}|\boldsymbol{x}^{0})\Bigr]
\end{equation}
In practice, computing ELBO is very time-consuming, because of the large number of denoising steps. So in practical training, we sample an $ n $ uniformly from $ \{1,...,N\} $ and compute the loss:
\begin{equation}
    \mathcal{L}_n(\theta)=\mathbb{D}_{\mathrm{KL}}\Bigl[q(\boldsymbol{x}^{n-1}|\boldsymbol{x}^{n},\boldsymbol{x}^{0})||p_\theta(\boldsymbol{x}^{n-1}|\boldsymbol{x}^{n})\Bigr]
\end{equation}
The posterior $ q(\boldsymbol{x}^{n-1}|\boldsymbol{x}^{n},\boldsymbol{x}^{0}) $ can be represented as:
\begin{equation}
    q(\boldsymbol{x}^{n-1}|\boldsymbol{x}^{n},\boldsymbol{x}^{0})=\mathcal{N}\Bigl(\boldsymbol{x}^{n-1};\Tilde{\boldsymbol{\mu}}(\boldsymbol{x}^{n},\boldsymbol{x}^{0},n),\Tilde{\beta}_n\boldsymbol{I}\Bigr)
\end{equation}
where
\begin{equation}
    \Tilde{\boldsymbol{\mu}}(\boldsymbol{x}^{n},\boldsymbol{x}^{0},n)=\frac{\sqrt{\Bar{\alpha}_{n-1}}\beta_n}{1-\Bar{\alpha}_{n-1}}\boldsymbol{x}^{0}+\frac{\sqrt{\alpha_n}(1-\Bar{\alpha}_{n-1})}{1-\Bar{\alpha}_{n}}\boldsymbol{x}^{n}
\end{equation}
\begin{equation}
    \Tilde{\beta}_n=\frac{1-\Bar{\alpha}_{n-1}}{1-\Bar{\alpha}_{n}}\beta_n
\end{equation}
\mbox{\cite{Ho}} proposes to speed up the training of DDPMs by simplifying the loss function:
\begin{equation}
\label{vanilla_loss}
    \mathcal{L}_n^{\mathrm{simple}}(\theta)=\mathbb{E}_{\boldsymbol{x}^0,n,\epsilon}\biggl[\Bigl\lVert \epsilon-\epsilon_\theta(\sqrt{\Bar{\alpha}_n}\boldsymbol{x}^0+\sqrt{1-\Bar{\alpha}_n}\epsilon) \Bigr\rVert^2\biggr]
\end{equation}
where $ \epsilon_\theta(\cdot) $ is a neural network used to predict the reparameterization noise $ \epsilon $ (see Eq. \mbox{\ref{forward}}). At each iteration of training, we draw $ n\sim \mathrm{Uniform}(1,...,N)$ and $ \epsilon\sim \mathcal{N}(0,1)$, and take the gradients on $ \nabla_\theta L_n^{\mathrm{simple}}(\theta) $. With the help of the noise prediction network $ \epsilon_\theta(\cdot) $,  we can recover the data at the $(n-1)$-th step given $ \boldsymbol{x}^n $ by drawing:   
\begin{equation}
\label{recover}
    \boldsymbol{x}^{n-1}\sim \mathcal{N}\biggl(\frac{1}{\sqrt{\alpha_n}}\Bigl(\boldsymbol{x}^n-\frac{1-\alpha_n}{\sqrt{1-\Bar{\alpha}_n}}\epsilon_\theta(\boldsymbol{x}^n,n)\Bigr),\frac{1-\Bar{\alpha}_{n-1}}{1-\Bar{\alpha}_{n}}\beta_nI\biggr)
\end{equation}
 Please refer to~\mbox{\cite{Ho}} for more details on DDPMs.

\subsection{Marked Temporal Point Processes}
Marked Temporal Point Processes (MTPPs) have been widely used to model discrete events in continuous time. A realization of an MTPP can be formulated as an event sequence $ \mathcal{S}=\{(m_i,t_i)\}_{i=1}^L $, where each event is  tagged with a mark $m_i$ and an arrival time $t_i$. The event marks contain the semantics of each event. Following previous works, we define them to be the labels of event types for simplicity, i.e., $ m_i=1,2,...,K $, where $K$ is the total number of event types. 

Most existing works on TPP model predict the mark and time of the next event given an event sequence as history, which reduces to the conditional joint probability $p(m_j,t_j|H_{t_{j-1}})$, where $ H_{t_{j-1}} $ refers to all history events up to the time $ t_{j-1} $. The target of TPP modelling is to fit a parameterized function:
\begin{equation}
g_{\boldsymbol{\Theta}}(m,t;H_{t'})=p(m,t|H_{t'};\boldsymbol{\Theta})
\end{equation}
With this network, given any history $ H_{t'} $, our model should be able to probabilistically predict the next potential event. There are various approaches to parameterize $ g_{\boldsymbol{\Theta}}(m,t;H_{t'}) $. Conventional TPP models propose to model intensity functions $ \lambda_{m}(t|H_{t'};\boldsymbol{\Theta}) $, and obtain the pdf by:
\begin{equation}
    p(m,t|H_{t'};\boldsymbol{\Theta})=\log \lambda_{m}(t|H_{t'};\boldsymbol{\Theta})-\int_{t'}^t\lambda(s|H_{t'};\boldsymbol{\Theta})ds
\end{equation}
where $\lambda(t|H_{t'};\boldsymbol{\Theta})=\sum_{k=1}^K \lambda_k(t|H_{t'};\boldsymbol{\Theta})$ is the total intensity of all event marks. The intensity function can be parameterized by either simple function forms or neural networks. On the other hand, some intensity-free TPP models~\mbox{\cite{lognormmix,ctpp}} have been proposed, which model the pdf $ p(m,t|H_{t'};\boldsymbol{\Theta}) $ directly using techniques such as mixture distributions.
$ \boldsymbol{\Theta} $ stands for the trainable parameters learned by maximizing the log-likelihood of each training event sequence:
\begin{equation}
\mathcal{L}(\mathcal{S},\boldsymbol{\Theta})=\sum_{i=1}^L \log p (m_i,t_i|H_{t_{i-1}};\boldsymbol{\Theta})
\end{equation}
Most existing TPP models focus on one-step-ahead event prediction. However, the prediction of long-term events has become an essential task in practice. Previous works on long-horizon TPP modelling~\mbox{\cite{long,hypro}} formulate the problem as predicting future event sequences in a time horizon $ T' $ given history event sequences within a time span of $ T $. However, the number of events that fall in the prediction horizon is not fixed and is unknown before the prediction. To fulfil this problem setting, prediction needs to be rolled out on a large number of steps and cut off the events that exceed the time boundary, leading to a drop in evaluation efficiency and a waste of information. Therefore, instead, we formulate our problem as event prediction within a step limit. Specifically, given the event history of $ L $ steps, we predict future events of $ L' $ steps, i.e., $ \mathcal{S}'=\{(m_{L+j},t_{L+j})\}_{j=1}^{L'} $. The modelling target of our formulation is a conditional probability distribution $ p(\mathcal{S}'|\mathcal{S}) $.

\section{Denoising Diffusion with Event Sequences}
We formulate the framework of the denoising diffusion method on event sequences (see Fig.~\mbox{\ref{fig:diffusion}}). Event sequences do not lie in an Euclidean space, where vanilla DDPMs~\mbox{\cite{Ho}} originally work. Therefore, transformations are needed to map event sequences into a vector space so that the forward and backward processes can be carried out.
\subsection{Forward Process}
\label{forward-process}

\begin{figure*}[!t]
\centering
\includegraphics[width=6.5in]{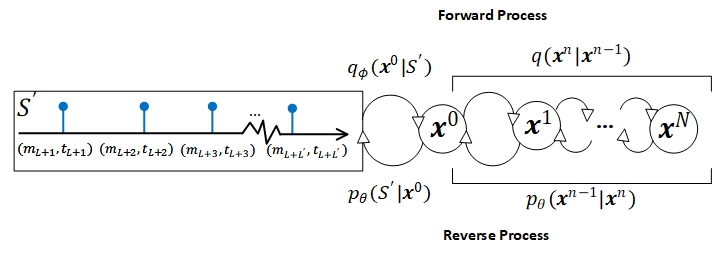}
\caption{Diffusion with event sequences. The target event sequence is mapped to the Euclidean space where the standard forward process is carried out to add Gaussian noise. The standard reverse process is followed by an inverse transformation that recovers the event sequence from its Euclidean embedding.}
\label{fig:diffusion}
\end{figure*}
The forward process, as mentioned in subsection \mbox{\ref{ddpm}}, refers to the construction of a noise-adding Markov chain, i.e., the joint probability $ q(\boldsymbol{x}^{1:N}|\boldsymbol{x}^0) $. However, in the case of TPP modelling, each data item is an event sequence $ \mathcal{S}' $, which is a mixture of positive continuous increasing variables $ t_{L+j} $ and discrete variables $ m_{L+j} $, thus the forward process of vanilla DDPM cannot be directly applied. To solve this problem,  inspired by~\mbox{\cite{diffusion-LM}}, as illustrated in the upper part of Fig.~\mbox{\ref{fig:diffusion}}, we propose to map the target event sequence $ \mathcal{S}' $ to a continuous real space. For this purpose, we define a probability $ q_\phi(\boldsymbol{x}^0|\mathcal{S}') $, where $ \boldsymbol{x}^0=[\boldsymbol{x^0_1;...;x^0_j;...;x^0_{L'}}]\in \mathbb{R}^{L'(d+1)} $. Each $ \boldsymbol{x}^0_j\in \mathbb{R}^{d+1} $ is composed of a time component $ x^{\mathrm{time}}_j\in \mathbb{R} $ and a mark component $ \boldsymbol{x}^{\mathrm{mark}}_j\in \mathbb{R}^{d} $:
\begin{equation} 
\label{x0}
    \boldsymbol{x}^0_j=[x^{\mathrm{time}}_j;\boldsymbol{x}^{\mathrm{mark}}_j]
\end{equation}
The mapping probability can then be factorized as:
\begin{equation}
    q_\phi(\boldsymbol{x}^0|\mathcal{S}')=q(\boldsymbol{x}^{\mathrm{time}}|\boldsymbol{t})q_\phi(\boldsymbol{x}^{\mathrm{mark}}|\boldsymbol{m})
\end{equation}
where $ \boldsymbol{x}^{\mathrm{time}}=[x^{\mathrm{time}}_1,...,x^{\mathrm{time}}_j,...,x^{\mathrm{time}}_{L'}]^{\mathsf{T}}\in \mathbb{R}^{L'} $ and $ \boldsymbol{x}^{\mathrm{mark}}=[\boldsymbol{x}^{\mathrm{mark}}_1;...;\boldsymbol{x}^{\mathrm{mark}}_j;...;$ $\boldsymbol{x}^{\mathrm{mark}}_{L'}]\in \mathbb{R}^{L'd} $ are the concatenation of all time and mark components respectively. $ \boldsymbol{t}=[t_{L+1}, t_{L+2},..., $ $t_{L+L'}]^{\mathsf{T}} $ is the vector of arrival times and $ \boldsymbol{m}=[m_{L+1}, m_{L+2},..., m_{L+L'}]^{\mathsf{T}} $ is an integer vector of event marks.

\begin{figure}[!t]
    \centering
    \includegraphics[width=2.5in]{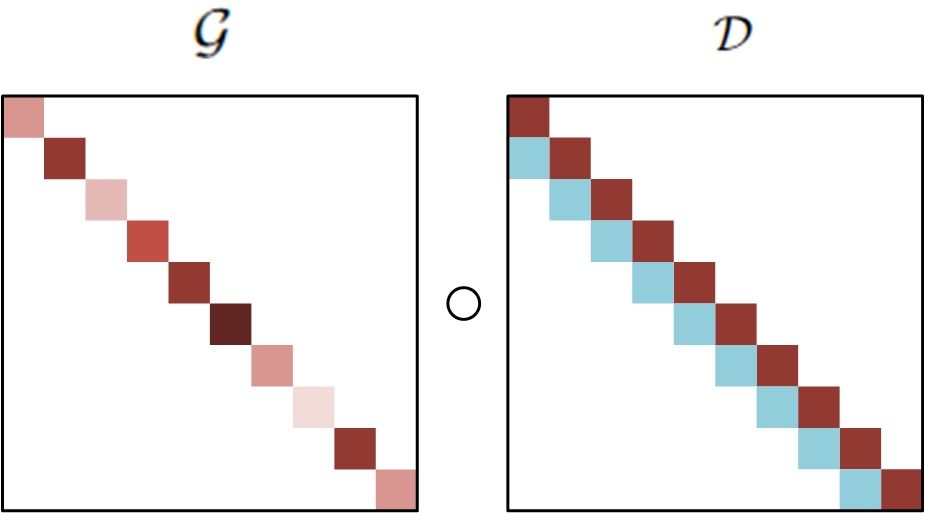}
    \caption{Jacobians of the time mapping transformations. $ \mathcal{D} $ is the bidiagonal difference matrix that computes the time intervals between events that are next to each other.  $ \mathcal{G} $ is an element-wise logarithm operation.}
    \label{fig:time_trans}
\end{figure}

The time mapping is defined to be a deterministic projection for simplicity, which makes the time mapping pdf a Dirac delta function:
\begin{equation}
    q(\boldsymbol{x}^{\mathrm{time}}|\boldsymbol{t})=\delta(\boldsymbol{x}^{\mathrm{time}}-\mathcal{M}\boldsymbol{t})
\end{equation}
where $ \mathcal{M}=\mathcal{G}\circ \mathcal{D} $ is an invertible transformation. 
The transformation $ \mathcal{D} $ is the pairwise difference operation, which has 1 on its diagonal, -1 along the diagonal below, and zeros elsewhere. It computes the time intervals between neighbouring events, i.e., $ \tau_i=t_i-t_{i-1} $. $ \mathcal{G} $ computes the elementwise logarithm, whose Jacobian only has non-zero elements on the diagonal, yielding $ \boldsymbol{x}^{\mathrm{time}}_i=\log \tau_i $. The Jacobians of $ \mathcal{G} $ and $ \mathcal{D} $ are shown in Fig.~\mbox{\ref{fig:time_trans}}. This transformation $ \mathcal{M} $ guarantees that $ \boldsymbol{x}^{\mathrm{time}} $ falls in the continuous real space, where the vanilla DDPM can be applied.

On the other hand, the mark mapping is defined as a parameterized Gaussian distribution:
\begin{equation}
    q_\phi(\boldsymbol{x}^{\mathrm{mark}}|\boldsymbol{m})=\mathcal{N}\Bigl(\boldsymbol{x}^{\mathrm{mark}};\mathrm{EMB_\phi(\boldsymbol{m})},\sigma_0I\Bigr) 
\end{equation}
where $ \mathrm{EMB_\phi} $ is an embedding layer with learnable parameters $ \phi $, and $ \sigma_0 $ is a hyperparameter that controls the variance. However, unlike time mapping, this mark mapping is probabilistic and not invertible, so we need to train another network $ \mathrm{EMB}^{-1}_\theta $ to recover the categorical mark distribution from $ \boldsymbol{x}^{\mathrm{mark}} $ to complete the reverse process (this will be discussed in the next subsection).

We then get the starting diffusion vector $ \boldsymbol{x}^0 $ by concatenating the time and mark component obtained above according to Eq.~\mbox{\ref{x0}}. The rest of the forward process follows the standard DDPM procedure. The whole forward process is then defined as the joint:
\begin{equation}
    \begin{split}
        q_\phi(\boldsymbol{x}^{0:N}|\mathcal{S}')&=q_\phi(\boldsymbol{x}^0|\mathcal{S}')q(\boldsymbol{x}^{1:N}|\boldsymbol{x}^0)\\
        &=q(\boldsymbol{x}^{\mathrm{time}}|\boldsymbol{t})q_\phi(\boldsymbol{x}^{\mathrm{mark}}|\boldsymbol{m})q(\boldsymbol{x}^{1:N}|\boldsymbol{x}^0)\\
    \end{split}
\end{equation}
where $ q(\boldsymbol{x}^{1:N}|\boldsymbol{x}^0) $ is formulated as the forward Markov chain given by Eq.~\mbox{\ref{forward2}}.

\subsection{Reverse Process}
\label{sub:reverse}
The vanilla diffusion model given by Eq.~\mbox{\ref{vanilla_loss}} uses the denoising network $ \epsilon_\theta $ to predict reparameterization noise $ \epsilon $. Similar to~\mbox{\cite{diffusion-LM,non-autoregressive}}, we find that using a denoising network $ f_\theta $ to predict $ \boldsymbol{x}^0 $ yields better results. In this case, a denoising step can be done by sampling:
\begin{equation}
\begin{split}
    \boldsymbol{x}^{n-1}\sim \mathcal{N}\Biggl(\frac{\sqrt{\alpha_n}(1-\Bar{\alpha}_{n-1})}{1-\Bar{\alpha}_{n}}\boldsymbol{x}^n+&\frac{\sqrt{\Bar{\alpha}_{n-1}}\beta_n}{1-\Bar{\alpha}_{n}}f_\theta(\boldsymbol{x}^n,n),\\
    &\frac{1-\Bar{\alpha}_{n-1}}{1-\Bar{\alpha}_{n}}\beta_nI\Biggr)
\end{split}
\label{denoising-step}
\end{equation}
The architecture of the denoising network will be discussed in the next section. However, different from vanilla DDPMs, as given by the lower part of Fig.~\mbox{\ref{fig:diffusion}}, we need an additional step to reconstruct the event sequences from $ \boldsymbol{x}^0 $, i.e.,
\begin{equation}
    p_\theta(\mathcal{S'}|\boldsymbol{x}^0)=p(\boldsymbol{t}|\boldsymbol{x}^{\mathrm{time}})p_\theta(\boldsymbol{m}|\boldsymbol{x}^{\mathrm{mark}})
\end{equation}
Similar to the forward process, the reconstruction of arrival times is deterministic, and since the forward transformation $ \mathcal{M} $ is invertible, the time reconstruction pdf is a Dirac delta function as follows:
\begin{equation}
    p(\boldsymbol{t}|\boldsymbol{x}^{\mathrm{time}})=\delta(\boldsymbol{t}-\mathcal{M}^{-1}\boldsymbol{x}^{\mathrm{time}})
\end{equation}
As mentioned in~\mbox{\ref{forward-process}}, a trainable network $ \mathrm{EMB}^{-1}_\theta $ is needed to reconstruct the probability space of event marks:
\begin{equation}
    p_\theta(\boldsymbol{m}|\boldsymbol{x}^{\mathrm{mark}})=\mathrm{Categorical}\Bigl(\mathrm{EMB}^{-1}_\theta(\boldsymbol{x}^{\mathrm{mark}})\Bigr)
\end{equation}
In practice, we formulate $ \mathrm{EMB}^{-1}_\theta $ as a Multi-Layer Perceptron (MLP) appended with a Softmax layer:

\begin{equation}
\begin{split}
\mathrm{EMB}&^{-1}_\theta(\boldsymbol{x}^{\mathrm{mark}})=\\
&\mathrm{Softmax}\Biggl[\boldsymbol{W}_2\biggl(\mathrm{GELU(\boldsymbol{W}_1\boldsymbol{x}^{\mathrm{mark}}+\boldsymbol{b}_1)}\biggr)+\boldsymbol{b}_2\Biggr]
\end{split}
\end{equation}
where $ \boldsymbol{W}_1 $, $ \boldsymbol{W}_2 $, $ \boldsymbol{b}_1 $, $ \boldsymbol{b}_2 $ are trainable parameters.
The whole reverse process now becomes:
\begin{equation}
\begin{split}
    p_\theta(\mathcal{S}',\boldsymbol{x}^{0:N})&=p_\theta(\boldsymbol{x}^{0:N})p_\theta(\mathcal{S'}|\boldsymbol{x}^0)\\
    &=p_\theta(\boldsymbol{x}^{0:N})p(\boldsymbol{t}|\boldsymbol{x}^{\mathrm{time}})p_\theta(\boldsymbol{m}|\boldsymbol{x}^{\mathrm{mark}})
\end{split}
\end{equation}
where $ p_\theta(\boldsymbol{x}^{0:N}) $ is formulated as the reverse Markov chain given by Eq.~\mbox{\ref{eq:reverse}}. 

Following~\mbox{\cite{diffusion-LM}}, we use the clamping trick $ 
\mathrm{Clamp}(\cdot) $ on the predicted $ \hat{\boldsymbol{x}}^0=f_\theta(\boldsymbol{x}^n,n) $ during sampling to improve prediction accuracy. Similar to Eq.~\mbox{\ref{x0}}, $ \hat{\boldsymbol{x}}^0 $ is composed as $ \hat{\boldsymbol{x}}^0_j=[\hat{x}^{\mathrm{time}}_j;\hat{\boldsymbol{x}}^{\mathrm{mark}}_j] $.  The trick is to clamp the mark component $ \hat{\boldsymbol{x}}^{\mathrm{mark}}_j $ to the nearest mark embedding defined by $ \mathrm{EMB}_\phi $. In this case, the denoising step given in Eq. \mbox{\ref{denoising-step}} becomes:
\begin{equation}
\begin{split}
    \boldsymbol{x}^{n-1}\sim &\mathcal{N}\Biggl(\frac{\sqrt{\alpha_n}(1-\Bar{\alpha}_{n-1})}{1-\Bar{\alpha}_{n}}\boldsymbol{x}^n+\\
    &\frac{\sqrt{\Bar{\alpha}_{n-1}}\beta_n}{1-\Bar{\alpha}_{n}}\cdot\mathrm{Clamp}(f_\theta(\boldsymbol{x}^n,n)),\frac{1-\Bar{\alpha}_{n-1}}{1-\Bar{\alpha}_{n}}\beta_nI\Biggr)
\end{split}
\label{denoising-step-clamp}
\end{equation}
However, using the clamping trick at the early stage of the reverse process results in bad sample quality, because the model has not figured out which event mark the sample should be moving towards. The clamping trick's starting point slightly affects the model's performance.

\subsection{End-to-end Training}
As presented in previous subsections, we can apply standard DDPMs to TPP modelling by mapping event sequences to a continuous space. However, extra loss terms need to be added to guarantee the reconstruction ability to recover event sequences from the embedding space. The loss function can be divided into four terms, i.e., $ \mathcal{L}(\theta,\phi)=\mathcal{L}_1(\theta,\phi)+\mathcal{L}_2(\theta)+\mathcal{L}_3(\phi)+\mathcal{L}_4(\theta,\phi) $.

The first term is the standard diffusion loss, which measures the square error between the real data $ \boldsymbol{x}^0 $ and the reconstructed data given by the denoising network.
\begin{equation}
\begin{split}
    \mathcal{L}_1(\theta,\phi)=&\mathbb{E}_{n}\biggr[\boldsymbol{1}_{n>1}(n)\mathbb{E}_{\boldsymbol{x}^0,\boldsymbol{x}^n}\Bigr[\lVert \boldsymbol{x}^0-f_\theta(\boldsymbol{x}^n,n) \rVert^2\Bigr]\\
    &+\boldsymbol{1}_{n=1}(n)\mathbb{E}_{\boldsymbol{x}^0,\boldsymbol{x}^1}\Bigr[\lVert f_\theta(\boldsymbol{x}^1,1)-\mathrm{EMB_\phi(\boldsymbol{m})}\rVert^2\Bigr]\biggr]
\end{split}
\end{equation}
where $ \boldsymbol{1}_A(\cdot) $ is the indicator function that takes value $ 1 $ only when the variable falls in set $ A $, and 0 otherwise. The second term is the reconstruction loss:
\begin{equation}
    \mathcal{L}_2(\theta)=\mathbb{E}_{\boldsymbol{x}^0}\Biggl[\mathrm{CrossEntropy}\biggl(\boldsymbol{m},\mathrm{EMB}^{-1}_\theta(\boldsymbol{x}^{\mathrm{mark}})\biggr)\Biggr]
\end{equation}
which ensures that the event marks can be recovered from the embedding space. The third term is an embedding regularization term:
\begin{equation}
    \mathcal{L}_3(\phi)=\mathbb{E}_{\boldsymbol{x}^0}\Bigl[\bigl\lVert \sqrt{\Bar{\alpha}_N}\boldsymbol{x}^0 \bigl\rVert^2\Bigr]
\end{equation}
Since the cross-entropy loss in $ \mathcal{L}_2 $ tends to push the mark embeddings as far as possible, $ \mathcal{L}_3 $ is needed to ensure that the corrupted distribution at the maximum level $ N $ is close enough to the standard Gaussian, which is essential to the sample quality. 

The last term $ \mathcal{L}_4(\theta,\phi) $ is a regularization term to alleviate the problem of overfitting, which is critical to the performance of our model. We implement this regularization through weight decay in our experiment.
All the expectations above are estimated by sampling $ n\sim \mathrm{Uniform}(1,...,N) $, $ \boldsymbol{x}^0\sim q_\phi(\boldsymbol{x}^0|\mathcal{S}') $ and $ \boldsymbol{x}^{1:N}\sim q(\boldsymbol{x}^{1:N}|\boldsymbol{x}^0) $. By taking gradients on $ \nabla_\theta \mathcal{L}(\theta,\phi) $, we are able to train our diffusion-based long-horizon temporal point process model.

\section{Architecture}
\begin{figure*}[ht!]
	\centering
	\includegraphics[width=0.8\textwidth]{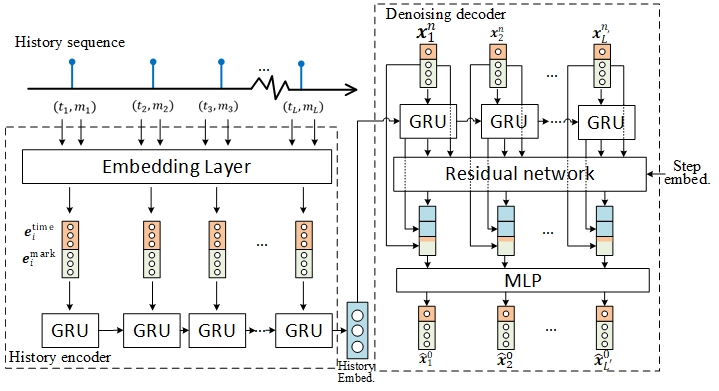}
	\vspace{-0.5em}
	\caption{Overview of the proposed denoising network for the diffusion model. The framework is comprised of a history encoder and a denoising decoder. The history encoder aggregates information from the known history event sequence, and outputs a history embedding, which will be used by the denoising decoder as a condition. The denoising decoder predicts $ \hat{\boldsymbol{x}}^0 $, which is used in the reverse process for sampling.}
	\centering
	\label{fig:framework}
	\vspace{-1em}
\end{figure*}

In this section, we illustrate the architecture of our denoising network, which is mainly composed of three parts, as shown in Fig.~\mbox{\ref{fig:framework}}, namely the history encoder, the sequential network, and the residual network.
\subsection{History Encoder} 
To condition our generation on known history events, we first use a history encoder to aggregate history information. The encoder is comprised of an embedding layer, which maps each event tuples into vectors, and an RNN layer that encodes the events sequentially.

\paragraph{Embedding layer}
Given a history event sequence $ \mathcal{S} $, the embedding layer maps each event, characterized as a tuple of the event mark $m_i$ and arrival time $t_i$, into a vector $ \boldsymbol{e}_i=[\boldsymbol{e}_i^{\mathrm{mark}};\boldsymbol{e}_i^{\mathrm{time}}]\in \mathbb{R}^{2d} $, i.e., a concatenation of mark embedding and time embedding. The mark embedding $ \boldsymbol{e}_i^{\mathrm{mark}}\in\mathbb{R}^d $ is obtained through a trainable discrete embedding module, which assigns an embedding vector for each mark label.
\begin{equation}
\boldsymbol{e}_i^{\mathrm{mark}}=\mathrm{EMB}^{\mathrm{enc}}_\theta(m_i)
\end{equation}
To capture flexible temporal patterns, we follow~\mbox{\cite{time2vec}} to adopt a learnable trigonometric module for time embedding $ \boldsymbol{e}_i^{\mathrm{time}}\in\mathbb{R}^d $:
\begin{equation}
    \boldsymbol{e}_i^{\mathrm{time}}[j]=\cos(\omega_jt_i+\beta_j)
\end{equation}
$\omega_j$, $ \beta_j$  are trainable parameters and $j$ is the index of the embedding dimension. 

\paragraph{Sequential Layer}
The sequential layer takes all the history event embeddings $\{{\boldsymbol{e}_1,...,\boldsymbol{e}_L}\}$ in the temporal order and generates a fixed-size history embedding $ \boldsymbol{h}\in\mathbb{R}^d $ to aggregate history information, which will be used as a condition in prediction. We adopt the Gated Recurrent Unit (GRU) as the backbone of the sequential layer and take the output of the last step as the history embedding.

\subsection{Denoising Decoder}
The input to the denoising decoder is the corrupted sample at the $ n $-th level, i.e., $ \boldsymbol{x}^n=[\boldsymbol{x}_1^n;\boldsymbol{x}_2^n;...;\boldsymbol{x}_{L'}^n]\in\mathbb{R}^{L'(d+1)} $. Furthermore, the history embedding $ \boldsymbol{h} $ produced by the history encoder is taken as a condition. The purpose of this decoder is to predict $ \hat{\boldsymbol{x}}^0 $, which is an approximation of  the uncorrupted sample $ \boldsymbol{x}^0 $. As mentioned in subsection \mbox{\ref{sub:reverse}}, the predicted $ \hat{\boldsymbol{x}}^0 $ can be used in the reverse process to remove Gaussian noise and recover the data. However, unlike previous diffusion models \mbox{\cite{Ho}} that basically work on image data, our model focuses on the sequential and contextual features of event sequences. Thus, we combine a GRU-based sequential layer with a convolutional residual network to incorporate both sequential and contextual features.

\paragraph{Sequential Layer} We use another GRU as the sequential layer of the denoising decoder to capture the sequential feature, the input to which is the diffusion variable $ \boldsymbol{x}^n $, where $ L' $ is the prediction length, and each $ \boldsymbol{x}_i^n\in\mathbb{R}^{d+1} $ corresponds to the $i$-th predicted event. The output of the sequential network, indicated as $ \boldsymbol{g}^{n}=[\boldsymbol{g}^{n}_1,...,\boldsymbol{g}^{n}_{L'}] $, will be fed to the residual network. Note that to condition our denoising on history data, the history embedding $h$ is passed to the sequential layer as the initial hidden state of the GRU. However, to achieve better flexibility, this sequential decoding layer does not share parameters with the sequential encoding layer introduced in the last subsection.

\begin{figure}[h!]
\label{residual}
	\centering
    \vspace{-1em}
	\includegraphics[width=0.3\textwidth]{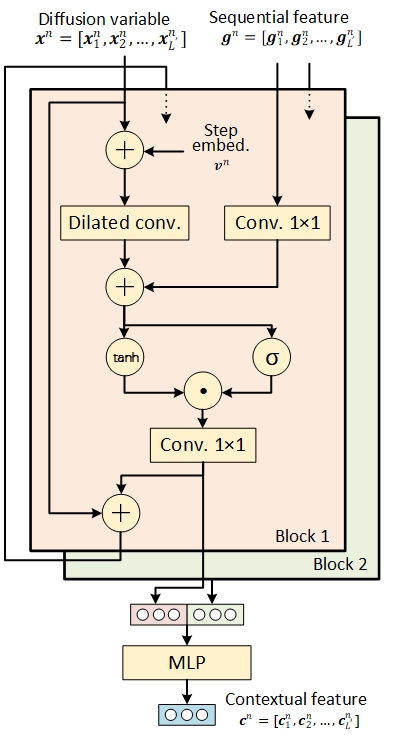}
	\vspace{-0.5em}
	\caption{Overview of a residual network with two blocks. Each residual block contains a dilated convolution layer and gated activation. The diffusion variable and sequential feature are the inputs to the network, and its output is the contextual feature. A diffusion step embedding is used internally to make the network aware of the noise level.}
	\centering
	\label{fig:residual}
	\vspace{-1em}
\end{figure}

\paragraph{Residual Network} We use a residual network with a similar architecture as~\mbox{\cite{TimeGrad,diffwave}}. It is made up of a stack of several residual blocks, each of which mainly contains a dilated convolution and a gated activation unit. The structure of the residual network is shown in Fig.~\mbox{\ref{fig:residual}}. 
The inputs to the residual network include the diffusion variable $ \boldsymbol{x}^n $, the output of the sequential network $ \boldsymbol{h}^{n} $ and the diffusion step embedding $ \boldsymbol{v}^n $. The step embedding is defined as an MLP transformation of a trigonometric positional embedding:
\begin{equation}
    \boldsymbol{v}^n=\boldsymbol{U}_2\Biggl(\mathrm{SiLU}\biggl(\boldsymbol{U}_1\boldsymbol{p}^n+\boldsymbol{c_1}\biggr)\Biggr)+\boldsymbol{c}_2
\end{equation}
$ \boldsymbol{U}_1 $,  $ \boldsymbol{U}_2 $,  $ \boldsymbol{c}_1 $,  $ \boldsymbol{c}_2 $ are trainable parameters, and $ \boldsymbol{p}^n\in\mathbb{R}^d $ is the trigonometric positional encoding.
The diffusion variable $ \boldsymbol{x}^n $ is first added up with the step embedding $ \boldsymbol{v}^n $ and then passed through a dilated convolutional layer to aggregate local contextual features. The sequential feature $ \boldsymbol{g}^{n} $ obtained by the sequential layer is added to the output of the dilated convolution after being transformed by a $ 1\times 1 $ convolution layer. A gated activation $ \tanh(\cdot)\odot \sigma(\cdot) $ proposed by~\mbox{\cite{diffwave}} is applied as the non-linear layer. The result of the activation goes on two branches, one of which is added with the diffusion input $ \boldsymbol{x}^n $ and passed to the subsequent residual block, and the other branch is the output of this residual block. The output of all blocks are concatenated and resized with an MLP to get the contextual feature.

The final output of the residual network is concatenated with $ \boldsymbol{x}^n $ and $ \boldsymbol{h}^{n} $ to produce a comprehensive feature that incorporates both sequential and contextual information (see Fig.~\mbox{\ref{fig:framework}}). Finally, an MLP is used to resize the feature to produce the predicted $ \hat{\boldsymbol{x}}^0 $.

\section{Experiments}
The proposed DLTPP model is evaluated on two measures, namely the Mean Absolute Error (MAE) of multi-step time prediction and the ACCuracy (ACC) of multi-step mark prediction. Experiments are performed on four datasets, including one synthetic dataset and three real-world datasets. Eight existing TPP models are chosen as our baselines to prove our model's superiority in long-term event prediction.

\subsection{Training Details}
We implement the proposed DLTPP model using the PyTorch framework. The sizes of hidden states and embedding vectors are set to 32 or 64. The number of layers of the residual network is tuned within 1, 2.  The maximum diffusion step $ N $ is set to 500. For all experiments, the length of the history event sequence is fixed to 20. During the process, the learning rate is set to 1e-3 initially and decays exponentially as training proceeds. All experiments are performed on a workstation equipped with an Intel Xeon Gold 6248R CPU, and NVIDIA Tesla T4 GPUs.

\subsection{Datasets}
\begin{figure*}[!t]
\centering
\subfloat[Number of event marks]{\includegraphics[width=0.32\linewidth]{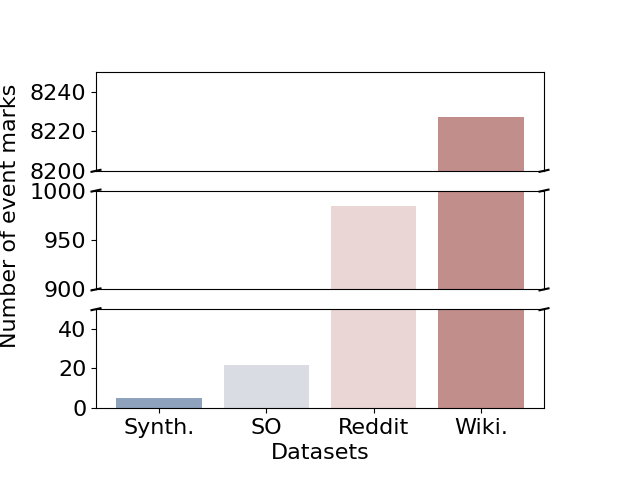}
}
\hfil
\subfloat[Number of sequences]{\includegraphics[width=0.32\linewidth]{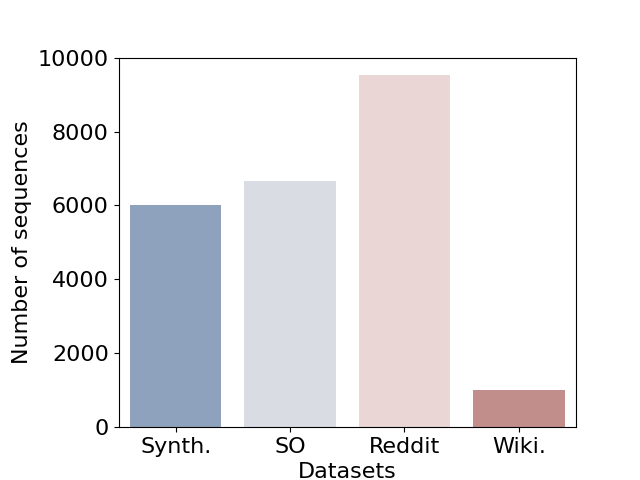}
}
\subfloat[Average sequence length]{\includegraphics[width=0.32\linewidth]{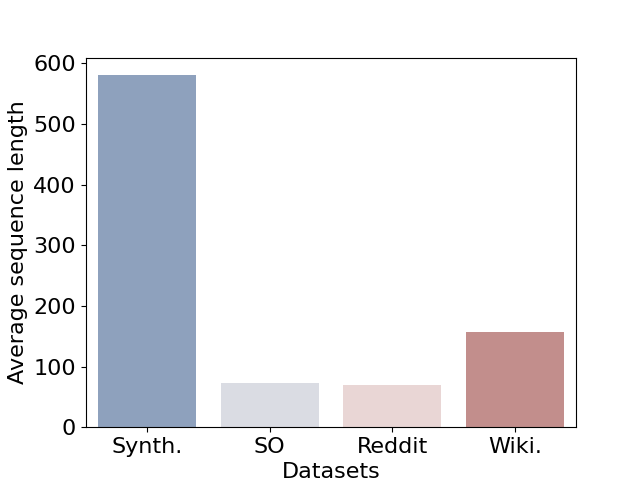}
}
\caption{Statistics of the four datasets (SO stands for StackOverflow). The datasets have different scales and characteristics.}
\label{fig:dataset}
\end{figure*}

We use 4 benchmark datasets to test our model's performance, including 1 synthetic dataset and 3 real-world datasets. As shown in Fig.~\mbox{\ref{fig:dataset}}, the four datasets differ from each other in statistical scales, which helps us prove the superiority of our model on datasets with different characteristics.  The synthetic dataset contains long sequences with few event marks. StackOverflow also has a small number of marks but has much shorter sequences. Reddit contains more event sequences and a much larger number of event marks. Wikipedia is a small dataset having an even larger number of event marks. For numerical stability, all event arrival times in the datasets are normalized to the range $ (0,100] $.






\subsubsection{Synthetic Dataset}
We use the same synthetic dataset as~\mbox{\cite{GNTPP}}. This dataset is generated using a Hawkes Process (HP), where the impact functions between each pair of the 5 event marks, i.e., $ g_{i,j}(t) $,  are uniformly-random chosen from 4 pre-defined kernel functions $ g_{a}(t) $, $ g_{b}(t) $, $ g_{c}(t) $, and $ g_{d}(t) $. Please refer to~\mbox{\cite{GNTPP}} for details of this synthetic dataset.

\subsubsection{Real-world Dataset}
\paragraph{StackOverflow~\mbox{\cite{stackoverflow}}} This dataset records the events of users earning badges on a question-and-answer website for programmers. There are 22 different marks in this dataset, characterizing 22 different badges.

\paragraph{Reddit~\mbox{\cite{reddit}}} This dataset contains the records of posts of some active users submitted to specific subreddits on a social media named Reddit. The 984 recorded subreddits are used as marks, and each event sequence corresponds to the posts of one active user.

\paragraph{Wikipedia~\mbox{\cite{reddit}}} This dataset contains the sequences of editing history of Wikipedia pages made by 8227 different users, which are characterized as 8227 different marks. Using this dataset helps us prove the superiority of our model even when the number of marks is large.

\subsubsection{Baselines}
\paragraph{RMTPP~\mbox{\cite{rmtpp}}} RMTPP is an early work on neural TPPs. It proposes an RNN-based autoregressive model to encode history events and predict the intensity function of the next event.

\paragraph{NHP~\mbox{\cite{nhp}}} NHP shares a similar structure with RMTPP, but applies a novel Continuous-time RNN, which is more flexible in encoding temporal contexts.

\paragraph{LogNormMix~\mbox{\cite{lognormmix}}} LogNormMix is an intensity-free TPP model. Instead of modelling the intensity function, it models the pdf of the next event directly using a mixture of multiple Log-Normal distributions.


\paragraph{TriTPP~\mbox{\cite{tritpp}}} TriTPP is a long-range TPP sequence generation model based on normalizing flows. Its generation is not conditioned on history events, and also does not consider event marks.

\paragraph{HYPRO~\mbox{\cite{hypro}}} HYPRO is a recently published non-autoregressive model dedicated to long-horizon event prediction, which adopts a generator-discriminator structure.

\paragraph{GNTPP~\mbox{\cite{GNTPP}}} GNTPP is a recently published autoregressive generative TPP model. It has multiple versions with different generative backbones. We choose the VAE-based (GNTPP-VAE) and diffusion-based (GNTPP-diffusion) versions for experimental comparison.

\paragraph{CTPP~\mbox{\cite{ctpp}}} CTPP is a recently published method that proposes to combine a continuous-time convolutional layer with an RNN network to incorporate global and local event features. We use their probabilistic decoder for time sampling in our experiment for fair comparison.

\subsection{Results on Multi-step Prediction}
\label{10-step}
We evaluate the proposed autoregressive DLTPP model using the time MAE and mark ACC of multi-step event prediction. Each batch of evaluation is conditioned on a history sequence of 20 events and predicts 10 subsequent events. The results are shown in Table~\mbox{\ref{tab:modelcomparision}}. The MAE shows the error of time prediction, and smaller values mean better performance. The ACC shows the accuracy of the mark prediction and larger values mean better performance. The best performance numbers on each dataset are bolded and the second-best ones are underlined. Note that TriTPP model does not depend on history events and does not consider event marks, thus it can not predict event marks and has bad time prediction performance.

\begin{table*}[ht!]
  \caption{Comparison of Results on multi-step-ahead Prediction} \label{tab:modelcomparision}
  \resizebox{2.00\columnwidth}{!}{
  \begin{tabular}{lrrrrrrrr}
    \toprule
    & \multicolumn{2}{c}{\texttt{Synthetic}}   & \multicolumn{2}{c}{\texttt{StackOverflow}} & \multicolumn{2}{c}{\texttt{Reddit}} & \multicolumn{2}{c}{\texttt{Wikipedia}}\\
    \cmidrule(lr){2-3} \cmidrule(lr){4-5} \cmidrule(lr){6-7} \cmidrule(lr){8-9}
    Methods & \multicolumn{1}{c}{MAE($\downarrow$)}      & \multicolumn{1}{c}{ACC($\uparrow$)}   & \multicolumn{1}{c}{MAE($\downarrow$)}          & \multicolumn{1}{c}{ACC($\uparrow$)}  & \multicolumn{1}{c}{MAE($\downarrow$)}      & \multicolumn{1}{c}{ACC($\uparrow$)} & \multicolumn{1}{c}{MAE($\downarrow$)}      & \multicolumn{1}{c}{ACC($\uparrow$)}\\
\midrule

RMTPP & 5.59 & 20.51 & 32.25 & \underline{35.13}& 42.18& 36.47 & \underline{12.51}& \underline{3.97}\\
NHP & \textbf{4.54} & 20.50 & 46.52 & 31.42  & 107.57 & 3.62 & 61.78 & 0.08\\
LogNormMix & 6.29 & 20.54 & 34.06 & 34.43  & 58.32 & 26.14 & 25.13 & 0.09\\
TriTPP & 386.78 & \multicolumn{1}{c}{-} & 311.50 & \multicolumn{1}{c}{-}  & 375.72 & \multicolumn{1}{c}{-} & 389.97 & \multicolumn{1}{c}{-}\\
HYPRO & 5.07 & 20.50 & 32.26 & 34.48  & 57.87 & \underline{49.84} & 13.13& 3.73\\
GNTPP-diffusion & 4.76 & \underline{20.55} & 37.24 & 34.50  & 95.35 & 26.94 & 27.00 & 0.20\\
GNTPP-VAE & 5.59 & 20.43 & \underline{25.85} & 34.79  & \underline{40.16} & 26.92 & 12.80 & 0.10\\
CTPP & 7.87& 20.49& 33.84& 33.83& 58.17& 34.91& 28.83& 1.43\\
\midrule
DLTPP & \underline{4.65} & \textbf{24.76} & \textbf{24.07} & \textbf{50.35}& \textbf{37.80}& \textbf{56.66}& \textbf{10.92}& \textbf{24.07}\\


\bottomrule
\end{tabular}
    }
\end{table*}

Our proposed model has the best mark prediction performance on all 4 datasets and surpasses the second-best model by a relatively large margin. This is probably because our non-autoregressive model is less affected by error accumulation, which is a major problem of autoregressive models. However, while HYPRO is non-autoregressive in general, it heavily relies on its autoregressive generator for sequence generation, which also leads to performance degradation.

As for time prediction results, our model has the best performance on three of the datasets and ranks second on the synthetic dataset. The explanation is that error accumulation has less effect on arrival times, which are continuous one-dimensional values while having a large effect on discrete event marks. Although our model does not beat all baseline models on all 4 datasets in time prediction, it poses the best consistency, since although NHP has a slightly better MAE result on the synthetic dataset, it performs quite poorly on the other three datasets.







\begin{figure}[!ht]
\centering
\subfloat[StackOverflow]{\includegraphics[width=0.45\linewidth]{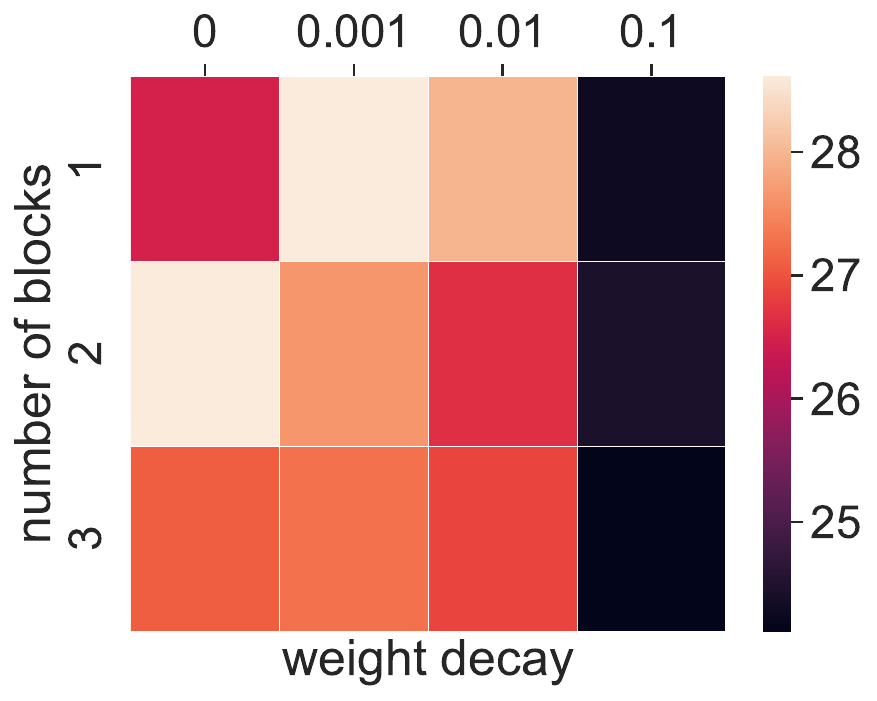}
}
\hfil
\subfloat[Reddit]{\includegraphics[width=0.45\linewidth]{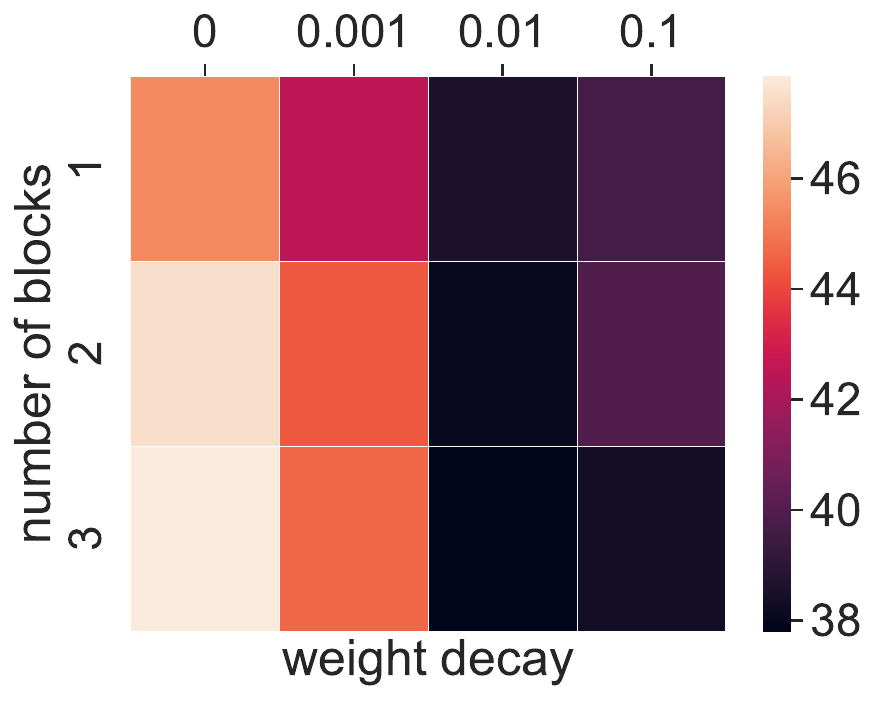}
}
\caption{MAE performance of DLTPP with different combinations of block numbers and weight decay.}
\label{fig:param_mae}
\end{figure}

\begin{figure}[!ht]
\centering
\subfloat[StackOverflow]{\includegraphics[width=0.45\linewidth]{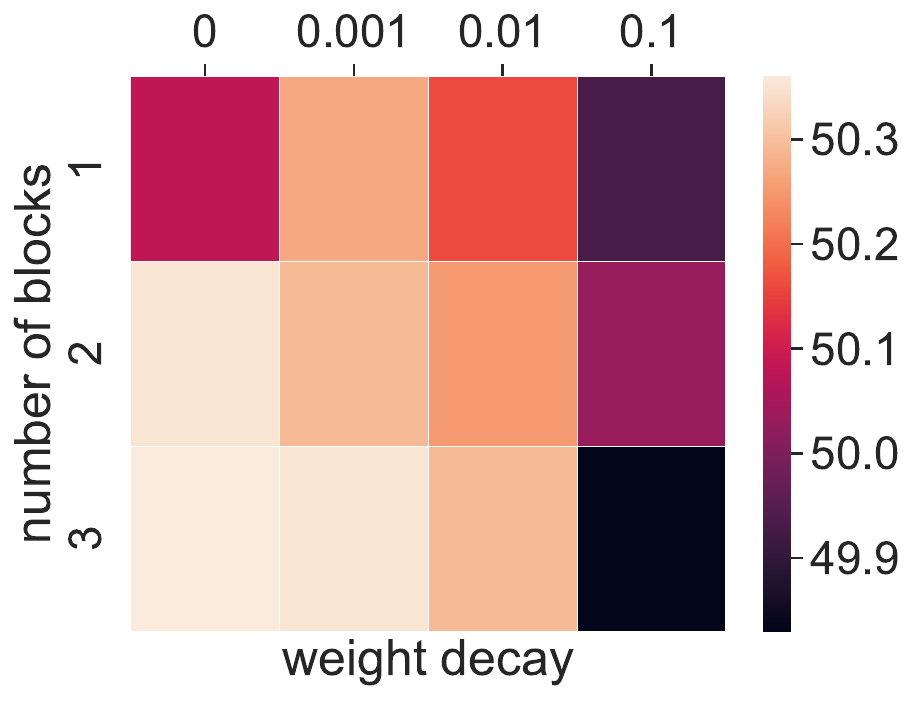}
}
\hfil
\subfloat[Reddit]{\includegraphics[width=0.45\linewidth]{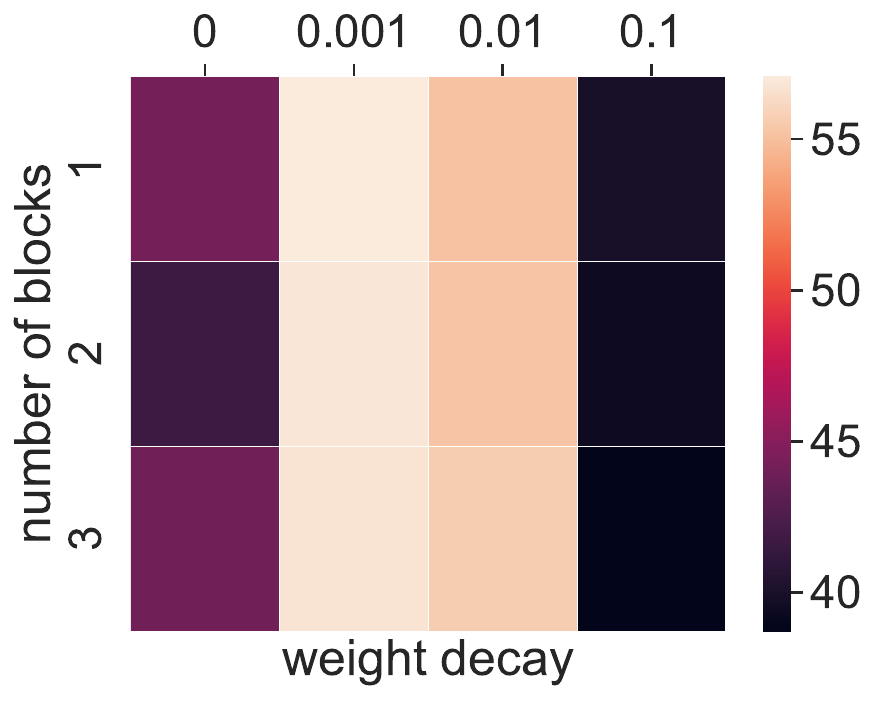}
}
\caption{ACC performance of DLTPP with different combinations of block numbers and weight decay (\%).}
\label{fig:param_acc}
\end{figure}

\subsection{Parameter Analysis}

We observe that the model performance is sensitive to the number of residual blocks of the residual network, as well as the coefficient of the regularization term. To explore the influence of these two hyperparameters on the performance of our model, we tune the number of blocks within [1, 2, 3] and the weight decay parameter within [0, 0.001, 0.01, 0.1]. In Fig.\mbox{\ref{fig:param_mae}}, we show the MAE results of our model trained with different combinations of these two parameters. We find that different datasets exhibit different preferences for the parameters. On StackOverflow, the model performs best when the weight decay is set to 0.1, and more residual blocks tend to enhance the prediction quality. However, the best result on Reddit is achieved by the model with 3 residual blocks when weight decay is set to 0.01. 

The ACC results of different parameter combinations are shown in Fig.~\mbox{\ref{fig:param_acc}}. On StackOverflow, the model's ACC performance gets better when using a smaller weight decay coefficient and more residual blocks, but the performance gaps between different parameter combinations are very small. On Reddit, however, the ACC performance is much more competitive when the weight decay is set to 0.001, and is not quite sensitive to the number of blocks. Generally speaking, using multiple residual blocks tends to yield better prediction performance than using only one block, and the best choice of weight decay varies with different datasets, thus requiring careful tuning. 

\subsection{Ablation Study}



\begin{figure}[!ht]
\vspace{-2em}
\centering
\subfloat[RNN-based]{\includegraphics[width=0.48\linewidth]{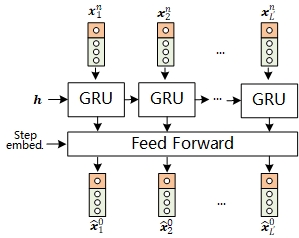}
\label{rnn}
}
\hfil
\subfloat[MLP-based]{\includegraphics[width=0.48\linewidth]{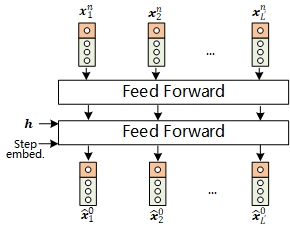}
\label{mlp}
}
\caption{Structures of two variants of the denoising decoders. The RNN-based network only captures sequential features while ignoring contextual features. The MLP-based network ignores both sequential and contextual features.}
\label{fig:variants}
\end{figure}

Our proposed denoising network incorporates both sequential and contextual information for better prediction performance.  To testify to the superiority of our design of the denoising network, we compare the MAE and ACC performance of our model against two of its simplified variants. In Fig.~\mbox{\ref{fig:variants}}, we show the structure of the denoising decoders of the two simplified variants. The first one uses a pure RNN-based network, which only captures sequential features while ignoring contextual features. The second simplifies the RNN to a feedforward layer, which captures neither sequential nor contextual features.

\begin{figure}[!ht]
\vspace{-2em}
\centering
\subfloat[MAE($\downarrow$]{\includegraphics[width=0.48\linewidth]{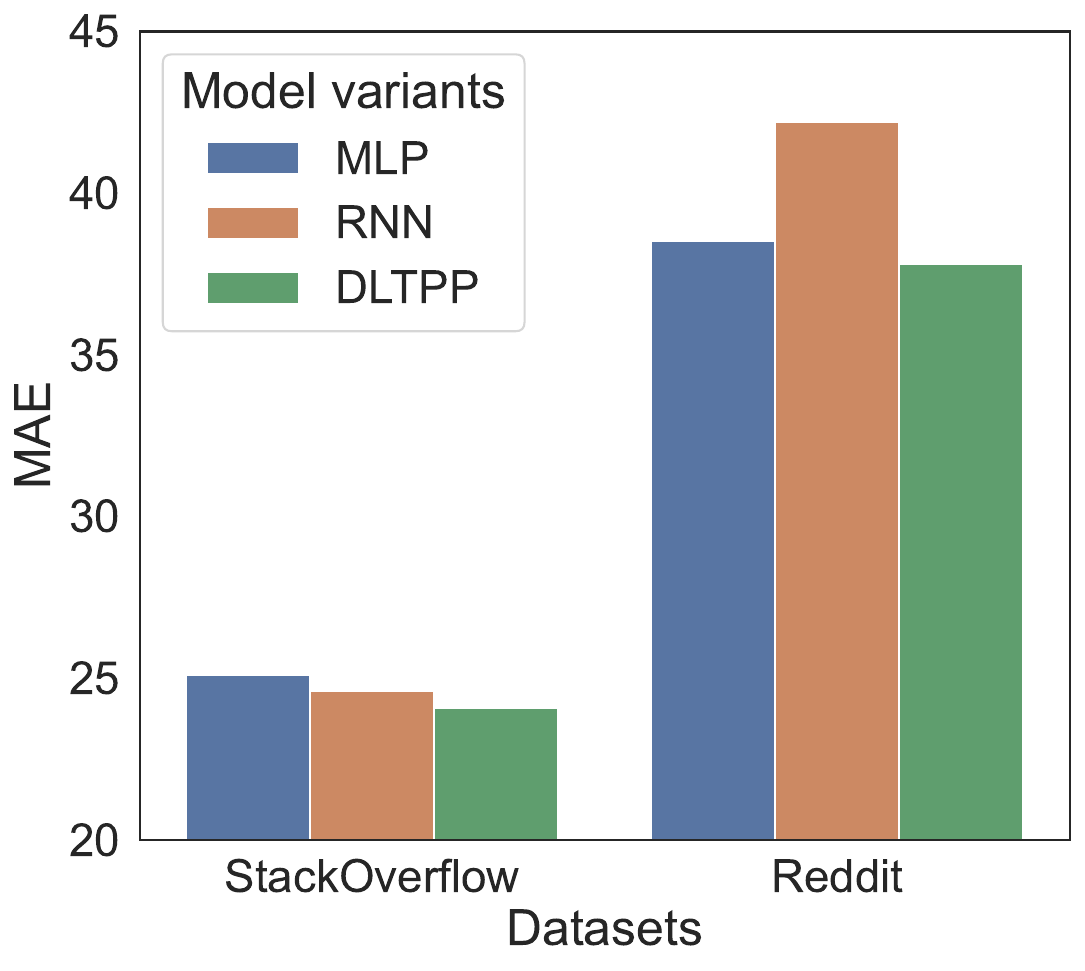}
\label{ablation_mae}
}
\hfil
\subfloat[ACC($\uparrow$)]{\includegraphics[width=0.48\linewidth]{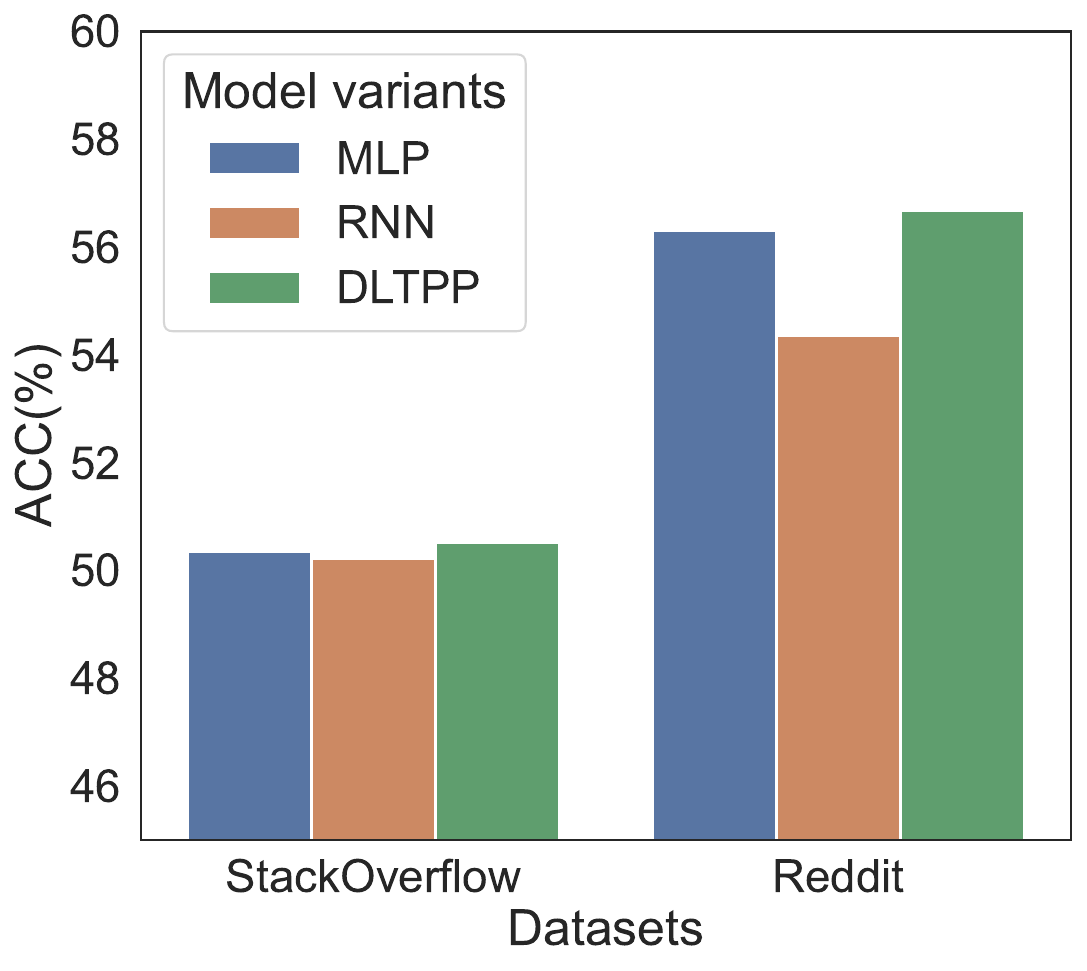}
\label{ablation_acc}
}
\caption{Comparison of MAE result between the proposed DLTPP and its simplified variants DLTPP-RNN and DLTPP-MLP.}
\label{fig:ablation}
\end{figure}

Fig.\mbox{\ref{fig:ablation}} shows the MAE and ACC results of DLTPP using different denoising architectures on the StackOverflow and Reddit data sets. Our denoising structure has the best result of MAE and ACC on both datasets, because of its capability of incorporating sequential and contextual features. The pure RNN-based framework ranks second on StackOverflow but comes last on Reddit. This might be caused by dataset-related issues. The StackOverflow dataset probably displays strong sequential dependencies of events, so an RNN that captures sequential features helps improve prediction performance. On the contrary, Reddit might exhibit a weak sequential correlation. In that case, an RNN does not help to improve prediction performance, but makes the model harder to train.




\begin{figure}[!ht]
\vspace{-2em}
\centering
\subfloat[MAE per event($\downarrow$)]{\includegraphics[width=0.48\linewidth]{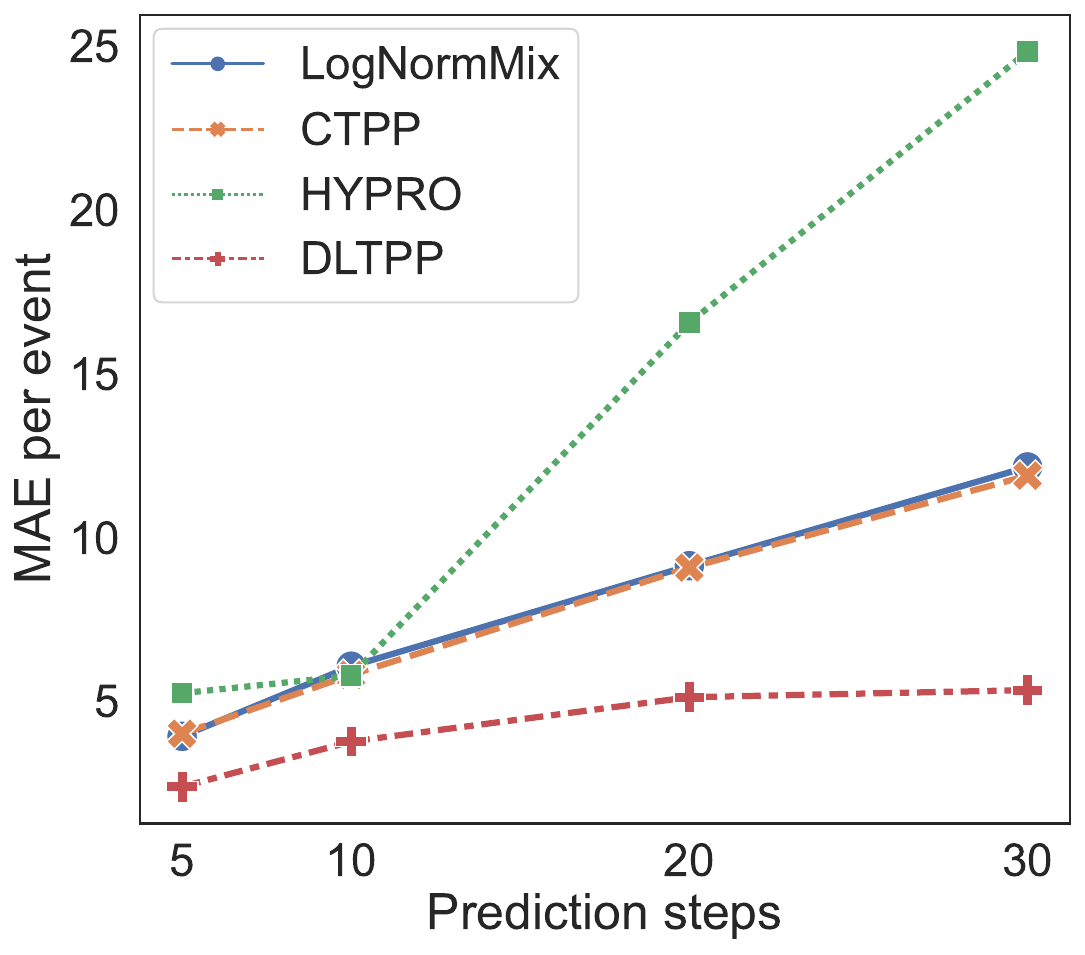}
\label{horizon_mae}
}
\hfil
\subfloat[ACC($\uparrow$)]{\includegraphics[width=0.48\linewidth]{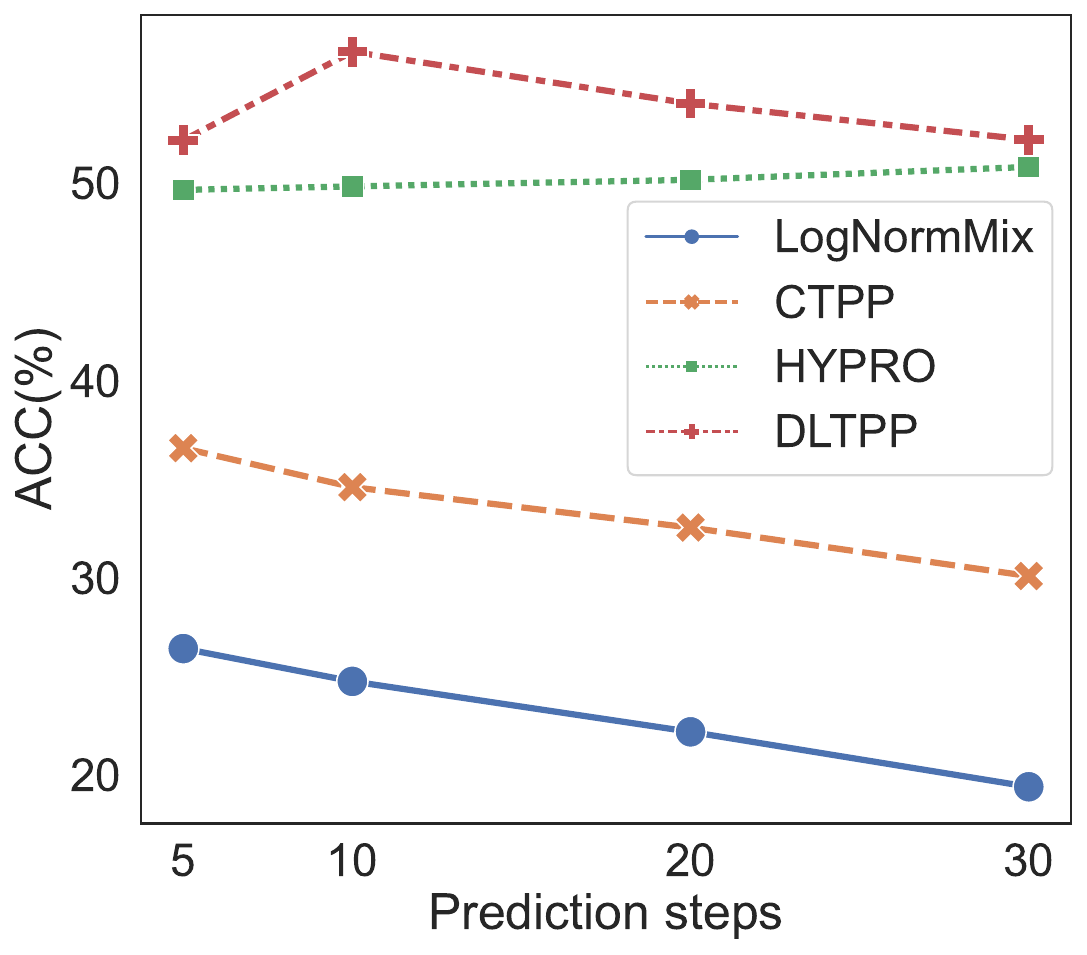}
\label{horizon_acc}
}
\caption{MAE and ACC results of long-term event prediction of different steps.}
\label{fig:horizon}
\end{figure}

Our model has consistently better results on both metrics with different prediction steps. The MAE per event of all models goes up as prediction steps increase, which is natural because more distant events are harder to predict, but the increasing slope of DLTPP is milder than other models. This is probably caused by DLTPP's ability to alleviate error accumulation. LogNorMix and CTPP are two typical autoregressive models whose MAE per event increases fast due to error accumulation. HYPRO is proposed as a long-horizon TPP model, but its time prediction performance deteriorates quickly as prediction steps increase, probably because the thinning algorithm that it uses to sample event times is less stable than sampling from closed-form distributions or generative models. However, HYPRO gives quite competitive ACC results, slightly below DLTPP. The reason for this is probably that HYPRO and DLTPP both adopt non-autoregressive structures, which makes wrong mark predictions less important to subsequent predictions. The ACC performance gap between DLTPP and HYPRO can be further explained by DLTPP's incorporation of sequential and contextual features. CTPP and LogNormMix present much lower mark prediction accuracy compared to non-autoregressive models, and the ACC decreases quickly as the number of time steps goes up as the result of error accumulation.

\section{Conclusion}
In this work, we propose a non-autoregressive diffusion-based temporal point process model for long-term event prediction in continuous time. We develop a way to perform denoising diffusion on event sequences through a bidirectional mapping mechanism and present a novel denoising network to incorporate sequential and contextual features. We provide thorough experimental results on long-term event prediction to validate the superiority of our model. For future work, we will extend our model to spatio-temporal event prediction and apply our model to more specific application scenarios.

\bibliographystyle{IEEEtran}
\bibliography{bibitems}

\end{document}